%% file: main.tex
\documentclass[conference]{IEEEtran}
\IEEEoverridecommandlockouts
\usepackage{cite}
\usepackage{amsmath,amssymb,amsfonts}
\usepackage{algorithmic}
\usepackage{graphicx}
\usepackage{textcomp}
\usepackage{amsmath}
\usepackage{breqn}
\usepackage{xcolor}
\def\BibTeX{{\rm B\kern-.05em{\sc i\kern-.025em b}\kern-.08em
    T\kern-.1667em\lower.7ex\hbox{E}\kern-.125emX}}
\begin{document}

\raggedbottom

\title{SimPO: Simultaneous Prediction and Optimization\\}

\author{
    \IEEEauthorblockN{Bing Zhang}
    \IEEEauthorblockA{IBM Research - Almaden \\
    San Jose, California \\
    bing.zhang@ibm.com}
    \and
    \IEEEauthorblockN{Yuya Jeremy Ong}
    \IEEEauthorblockA{IBM Research - Almaden \\
    San Jose, California \\
    yuyajong@ibm.com}
    \and
    \IEEEauthorblockN{Taiga Nakamura}
    \IEEEauthorblockA{IBM Research - Almaden \\
    San Jose, California \\
    taiga@us.ibm.com}
}


\maketitle

\begin{abstract}
Many machine learning (ML) models are integrated within the context of a larger system as part of a key component for decision making processes. Concretely, predictive models are often employed in estimating the parameters for the input values that are utilized for optimization models as isolated processes. Traditionally, the predictive models are built first, then the model outputs are used to generate decision values separately. However, it is often the case that the prediction values that are trained independently of the optimization process produce sub-optimal solutions. In this paper, we propose a formulation for the Simultaneous Prediction and Optimization (SimPO) framework. This framework introduces the use of a joint weighted loss of a decision-driven predictive ML model and an optimization objective function, which is optimized end-to-end directly through gradient-based methods.

\end{abstract}


\input{sections/01_introduction}
\input{sections/02_prior_art}
\input{sections/03_methodology}
\input{sections/04_conclusion}

\bibliographystyle{IEEEtran}
\bibliographystyle{chicago}
\bibliography{References.bib}
\vspace{12pt}

\end{document}

%% file: sections/01_introduction.tex
\section{Introduction}
The intersection between predictive machine learning (ML) algorithms within an optimization process has gained significant attention in recent years \cite{kotary2021end}. One such paradigm known as \textit{Predict, then Optimize}, entails the use of ML algorithms to estimate unknown parameters of a distribution, which is then subsequently used as inputs towards an optimization process \cite{stuckey2020dynamic, mandi2020interior}. The system is defined as a two-stage process where the predictive ML models are trained independently of the optimization problem. The predictive models are used as a method to estimate the unknown distribution, to which those output values are used as inputs towards an optimization process to maximize or minimize some objectives to arrive at an optimal decision.

\begin{figure}[]
\centerline{\includegraphics[scale=0.37]{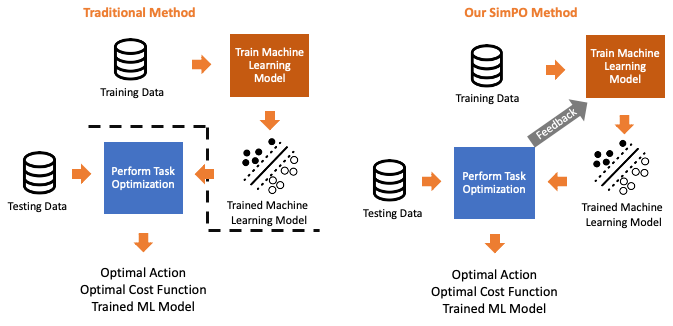}}
\caption{Process flow diagram between the \textit{Predict, then Optimize} versus the \textit{Simultaneous Prediction and Optimization} (SimPO) method.}
\label{Prior}
\end{figure}

One problem with such an approach is the two processes are performed independently from each other and are not considered as a single integrated process, as demonstrated in Figure \ref{Prior}. In the traditional process, the ML models are first trained against a loss function, then a separate optimization solver is applied to minimize or maximize a defined optimization objective. The key limitation of this approach is found in how the model parameters of the distribution are only optimized with respect to the loss function and not considering the end optimization task at hand. As a result, the predictions generated by the model results in an output which may influence the optimization process to produce a sub-optimal solution. 

Furthermore, another aspect that the ML models in this \textit{Predict, then Optimize} framework does not consider are the actions taken with respect to the input observations. In other words, the variable that is dependent on the input feature and its actions are not  considered jointly within the prediction of the model’s outcome. For example, for the problem optimizing the inventory of a particular product in a supermarket, the model which predicts the demand of the product would have to take into consideration the action as to whether to stock a certain quantity of a particular item. In this scenario, it may be possible that the demand of a particular item might be influenced by the potential quantity that the store may be looking to stock, therefore affecting its overall demand in the marketplace.

Hence, considering these two aspects, we propose a novel end-to-end joint process known as \textit{Simultaneous Prediction and Optimization} (SimPO), a method that considers a joint weighted function between the loss of prediction error and the optimization task loss. The framework relies on defining a joint loss function that combines both the ML model's loss function as well as the optimization cost function which its latent parameters can be jointly learned via stochastic gradient descent. This results in a model that maximizes the parameters of the relevant regions of the data distribution distribution which is based on the end optimization process, while also being able to derive the optimal solution to the optimization problem at hand. The proposed methodology can be widely applied to decision-making services and recommendation systems.

In the rest of this paper, the sections are organized as follows. Section 2 introduces related works around the intersection between ML and optimization processes. In Section 3, we outline the problem definition and the proposed methodology of our algorithm. Finally, Section 4 concludes the paper and also provides direction for future research directions.

%% file: sections/02_prior_art.tex
\section{Prior Art}
In this section, we describe some of the relevant prior art surrounding ML and optimization processes, and focusing our discussion around the \textit{Predict, then Optimize} framework. We further juxtapose some of the key differences between previous methods and our proposed method.

For the traditional two-stage method, typical ML algorithms that are used in the first stage of prediction include regression, neural networks, decision trees, and clustering models \cite{verwer2017auction,villarrubia2018artificial, abueidda2019prediction,wang2020ensemble}. These models are trained to maximize the objective values of the optimization problem, instead of including or minimizing the loss function. Wilder et al. propose a framework called decision-focused learning where the ML model is directly trained in conjunction with the optimization algorithm to produce high quality decisions \cite{wilder2019decision, wasserkrugensuring}. They take the discrete optimization problems into predictive models, which are typically trained via stochastic gradient descent.

Recently, many recent works in the ML and optimization space follow a paradigm called the \textit{Smart Predict, Then Optimize} (SPO) framework. The core idea of SPO framework is based on supervised learning, where a mapping from training inputs to outputs is learned. A surrogate loss function is derived from three upper bounds on SPO loss function \cite{elmachtoub2021smart}. Some studies extend this framework by providing two kinds of generalization bounds \cite{balghiti2019generalization}. However, the surrogate loss functions are not guaranteed to recover optimal decisions. Therefore, some other works start to make use of some special properties of the ML models, so that the models can be trained under the new loss without computing gradients. One example algorithm is the SPO decision tree method. As an extend idea, SPO forests is a methodology for training an ensemble of SPO decision tree to boost decision performance \cite{Elmachtoub2020decision}. 

Another research direction where the SPO framework is being applied is towards Reinforcement Learning (RL). Hu and Cheng propose a framework using neural networks and RL \cite{xinyi2020reinforcement} to solve combinatorial optimization problems. They train a recurrent neural network model to predict a distribution over different edges permutations given a directed graph. Using negative tour length as the reward signal, they optimize the parameters of the recurrent neural network using a policy gradient method. Another work studies the SPO framework in the context of sequential decision problems with missing parameters (formulated as Markov decision processes (MDPs)) that are solved via deep RL algorithms \cite{wang2021learning}.

In this work, instead of training the model based on the objective of the machine learning task at hand, we aim to learn a model based upon an end-to-end objective that the user is ultimately trying to accomplish. We utilize an end-to-end learning mechanism so that the ML process is directly predicted from raw inputs. Also highly related to our work are recent efforts in end-to-end policy learning, using value iteration effectively as an optimization procedure in similar networks, and multi-objective optimization \cite{donti2017task}. Given the similar motivations in modifying typically-optimized policies to address some tasks directly, we specifically focus on ML models of unknown variable structures and objective functions with action variables.

%% file: sections/03_methodology.tex
\section{Methodology}
\begin{figure*}[]
\centerline{\includegraphics[scale=0.78]{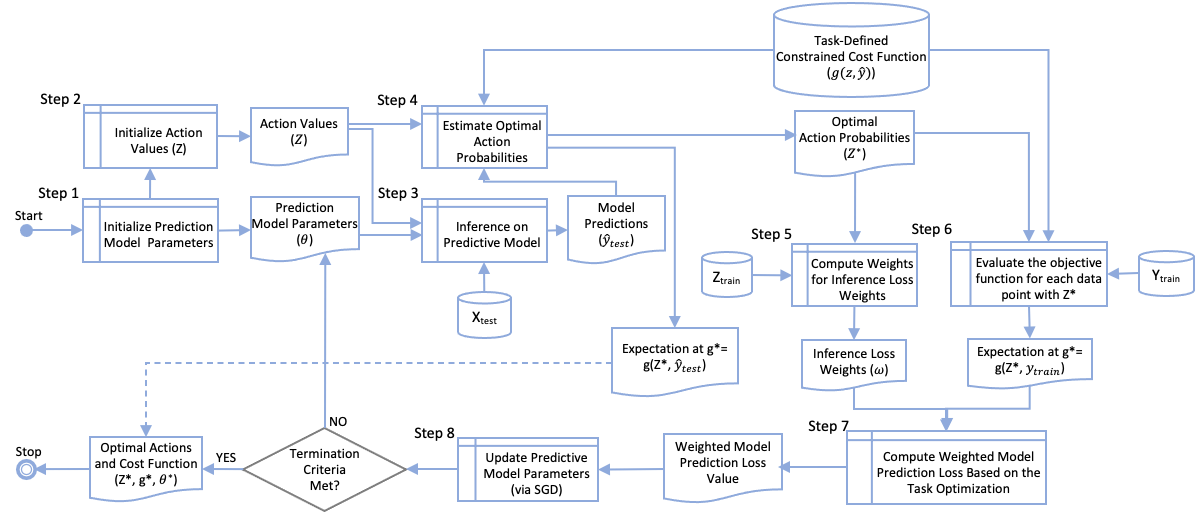}}
\caption{Flowchart of the proposed Simultaneous Prediction and Optimize Framework.}
\label{Methodology}
\end{figure*}

In this section, we introduce the SimPO problem definition and the methodology used to solve the problem, which is defined as a stochastic programming framework. Here, we introduce the notation used in this paper as well as the core framework, as shown in Figure \ref{Methodology}, which comprises of the different objective functions and constraints involved in the setup of the problem this framework aims to solve.

We consider a pair of input and output distribution of real-valued pair of values sampled from an unknown distribution, denoted as $(x \in X, y \in Y) \sim D$, which comprises of a pair of input features mapped to an output value. Furthermore, we define another variable for the action taken as denoted by $z \in Z$, which incurs a specific cost function defined as $L_{D}(z) = \mathbf{E}_{x,y\sim D}[f(x, y, z)]$. If the distribution of the data samples are known we can directly optimize for the best set of actions by minimizing the cost function directly, as $z^*_{D} = argmin_z \; L_{D}(z)$. Since we are not directly given the input and output pairs of the data distribution, a machine learning model is trained to estimate the corresponding parameters that map the input features, $X$, and output value, $Y$, pairs to the data distribution $D$. In the traditional SPO method this process attempts to model a conditional distribution using a parameterized model, as defined by $\hat{y} = Pr(y|x;\theta)$, which independently optimizes a loss function of the machine learning model defined by $argmin_{\theta} \; l(y, \hat{y})$, that minimizes the error between the ground truth output, $y$, and the output from the machine learning model, $\hat{y}$, parameterized by $\theta$. Subsequently, given a fixed model parametermized $L_{D}(z)$ is used to find the optimal set of action, $z^*$, given an unobserved input feature data sample, $x$, along with the output generated by the predictive model, $\hat{y}$. Here, the process of prediction and optimization are treated as an entirely disjoint processes.

We define the following joint weighted cost function as our objective function which considers both the weight of ML predictive loss and optimization function:
\begin{dmath}
    F(y_{train}, \hat{y}_{train}, y_{test}, \widetilde{z}, z^{*}_{train}, z^{*}_{test}) = l(y_{train}, \hat{y}_{train}) \times \omega(\widetilde{z}, z^{*}_{train}, \alpha) + g(\widetilde{z}, y_{test}) \times \gamma(z^{*}_{train}, z^{*}_{test}, \beta)
\end{dmath}

where $\widetilde{z} = [z_{min}, z_{max}]$ is the range of possible actions to consider, $l(y_{train}, \hat{y}_{train})$ is the predictive loss function, $\omega(\widetilde{z}, z^{*}_{train}, \alpha)$ is the weight for the predictive loss function, $g(z, y)$ is the task-constrained objective cost function that we want to minimize $argmin_{z} \; \mathbf{E}_{y \sim Pr(y|x,z,\theta)}[g(z,y)]$, and $\gamma(z^{*}_{train}, z^{*}_{test}, \beta)$ is the weight function for that. Furthermore, $\omega(\widetilde{z}, z^{*}_{train}, \alpha)$ is an increasing function with respect to the distance between $\widetilde{z}$ and $z^{*}_{train}$. $\gamma(z^{*}_{train}, z^{*}_{test}, \beta)$ is a decreasing function with respect to the distance between $z^{*}_{train}$ and $z^{*}_{test}$. We use the process outline in the flow diagram depicted in Figure \ref{Methodology} to optimize the objective function by and following steps:


\begin{enumerate}
    \item Given the initialized predictive model, $\hat{Y} = Pr(Y|X,Z;\theta)$, and $X_{test}$, we find the optimal set of action probabilities given the set of possible action ranges, $\widetilde{Z}$.
    \item Given the optimal action ranges and the historical label values, $Y_{train}$, we jointly compute the inference loss weights and the task optimization cost values.
    \item With the computed cost function and the weighted inference loss weights, we correspondingly compute the gradients and perform stochastic gradient descent to update the model weights with respect to the predictive model parameters.
    \item We repeat steps 1 to 3 until some termination criteria (according to specific problem) are met, for which we can determine our optimal action sets, $Z^{*}$, the optimal task-defined constrained cost function, $g^{*}(Z^{*},Y)$, and the optimal predictive model parameter $\theta^{*}$.
\end{enumerate}

%% file: sections/04_conclusion.tex
\section{Conclusion and Future Work}
In this work-in-progress paper, we presented a novel method for \textit{Simultaneous Prediction and Optimization} which jointly trains a ML model and optimization function based on a joint learning and optimization process. We demonstrated a plausible formulation and definition for how a joint weighted cost function considers both ML predictive loss function and the task-constrained optimization function. 
Given this formulation, our future work will entail performing experiments to demonstrate the viability of these methods on various synthetic datasets and real-world problems. In particular, one of the key objectives in the research aims to see if models that have sub-performing performance in certain regions of the distribution can be weighted in the optimization process that are much more relevant in the cost function input that yield better solutions.

%% file: main.bbl
\begin{thebibliography}{10}
\providecommand{\url}[1]{#1}
\csname url@samestyle\endcsname
\providecommand{\newblock}{\relax}
\providecommand{\bibinfo}[2]{#2}
\providecommand{\BIBentrySTDinterwordspacing}{\spaceskip=0pt\relax}
\providecommand{\BIBentryALTinterwordstretchfactor}{4}
\providecommand{\BIBentryALTinterwordspacing}{\spaceskip=\fontdimen2\font plus
\BIBentryALTinterwordstretchfactor\fontdimen3\font minus
  \fontdimen4\font\relax}
\providecommand{\BIBforeignlanguage}[2]{{%
\expandafter\ifx\csname l@#1\endcsname\relax
\typeout{** WARNING: IEEEtran.bst: No hyphenation pattern has been}%
\typeout{** loaded for the language `#1'. Using the pattern for}%
\typeout{** the default language instead.}%
\else
\language=\csname l@#1\endcsname
\fi
#2}}
\providecommand{\BIBdecl}{\relax}
\BIBdecl

\bibitem{kotary2021end}
J.~Kotary, F.~Fioretto, P.~Van~Hentenryck, and B.~Wilder, ``End-to-end
  constrained optimization learning: A survey,'' \emph{arXiv preprint
  arXiv:2103.16378}, 2021.

\bibitem{stuckey2020dynamic}
P.~J. Stuckey, T.~Guns, J.~Bailey, C.~Leckie, K.~Ramamohanarao, J.~Chan
  \emph{et~al.}, ``Dynamic programming for predict+ optimise,'' in
  \emph{Proceedings of the AAAI Conference on Artificial Intelligence},
  vol.~34, no.~02, 2020, pp. 1444--1451.

\bibitem{mandi2020interior}
J.~Mandi and T.~Guns, ``Interior point solving for lp-based prediction+
  optimisation,'' \emph{arXiv preprint arXiv:2010.13943}, 2020.

\bibitem{verwer2017auction}
S.~Verwer, Y.~Zhang, and Q.~C. Ye, ``Auction optimization using regression
  trees and linear models as integer programs,'' \emph{Artificial
  Intelligence}, vol. 244, pp. 368--395, 2017.

\bibitem{villarrubia2018artificial}
G.~Villarrubia, J.~F. De~Paz, P.~Chamoso, and F.~De~la Prieta, ``Artificial
  neural networks used in optimization problems,'' \emph{Neurocomputing}, vol.
  272, pp. 10--16, 2018.

\bibitem{abueidda2019prediction}
D.~W. Abueidda, M.~Almasri, R.~Ammourah, U.~Ravaioli, I.~M. Jasiuk, and N.~A.
  Sobh, ``Prediction and optimization of mechanical properties of composites
  using convolutional neural networks,'' \emph{Composite Structures}, vol. 227,
  p. 111264, 2019.

\bibitem{wang2020ensemble}
F.~Wang, Y.~Li, F.~Liao, and H.~Yan, ``An ensemble learning based prediction
  strategy for dynamic multi-objective optimization,'' \emph{Applied Soft
  Computing}, vol.~96, p. 106592, 2020.

\bibitem{wilder2019decision}
D.~B. . T.~M. Wilder, B., ``Melding the data-decisions pipeline:
  Decision-focused learning for combinatorial optimization,'' \emph{Proceedings
  of the AAAI Conference on Artificial Intelligence, 33(01), 1658-1665}, 2019.

\bibitem{wasserkrugensuring}
S.~Wasserkrug, O.~Davidovich, E.~Shindin, D.~Subramanian, P.~Ram, P.~Murali,
  D.~Phan, N.~Zhou, and L.~M. Nguyen, ``Ensuring the quality of optimization
  solutions in data generated optimization models.''

\bibitem{elmachtoub2021smart}
A.~N. Elmachtoub and P.~Grigas, ``Smart “predict, then optimize”,''
  \emph{Management Science}, 2021.

\bibitem{balghiti2019generalization}
O.~E. Balghiti, A.~N. Elmachtoub, P.~Grigas, and A.~Tewari, ``Generalization
  bounds in the predict-then-optimize framework,'' \emph{arXiv preprint
  arXiv:1905.11488}, 2019.

\bibitem{Elmachtoub2020decision}
J.~C. N.~L. A.~N.~Elmachtoub and R.~McNellis, ``Decision trees for
  decision-making under the predict-then-optimize framework,'' \emph{arXiv
  preprint arXiv:2003.00360}, 2020.

\bibitem{xinyi2020reinforcement}
H.~Xinyi and Y.~Cheng, ``Reinforcement learning for predict+ optimize,'' 2020.

\bibitem{wang2021learning}
K.~Wang, S.~Shah, H.~Chen, A.~Perrault, F.~Doshi-Velez, and M.~Tambe,
  ``Learning mdps from features: Predict-then-optimize for sequential decision
  problems by reinforcement learning,'' \emph{arXiv preprint arXiv:2106.03279},
  2021.

\bibitem{donti2017task}
P.~L. Donti, B.~Amos, and J.~Z. Kolter, ``Task-based end-to-end model learning
  in stochastic optimization,'' \emph{arXiv preprint arXiv:1703.04529}, 2017.

\end{thebibliography}
